\RequirePackage{fix-cm}
\documentclass[smallextended]{svjour3}       
\smartqed  
\usepackage{graphicx}
\usepackage{cite}
\usepackage{amsmath,amssymb,amsfonts}
\usepackage{algorithmic}
\usepackage{graphicx}
\usepackage{textcomp}
\usepackage{xcolor}
\usepackage{algorithm2e}
\usepackage{xcolor}

\SetAlFnt{\footnotesize}

\SetAlCapSty{xAlCapSty}

\SetCommentSty{xCommentSty}

\SetNlSty{mynlfont}{}{} 

\LinesNumbered

\SetSideCommentRight

\DontPrintSemicolon

\RestyleAlgo{algoruled}

\usepackage{hyperref}
\usepackage{mdframed}
\usepackage{lipsum}
\usepackage{amsmath}
\usepackage{amsfonts}
\usepackage[font=small]{caption}
\usepackage{ifpdf}
\usepackage{todonotes}
\usepackage{booktabs} 
\usepackage{url}
\usepackage{pifont}
\usepackage{color}
\usepackage{tcolorbox}
\usepackage{balance}
\usepackage{makecell}
\usepackage{verbatim}
\usepackage{tabularx}
\usepackage{colortbl}

\tcbuselibrary{breakable} 

\usepackage{algorithmic}
\usepackage{array}

\usepackage{bussproofs}
\usepackage{lipsum} 
\usepackage{mathpartir}

\usepackage{caption}
\usepackage{multirow}
\usepackage{float}
\usepackage{mwe}
\newcommand{\code}[1]{\texttt{#1}}

\usepackage{url}
\usepackage{todonotes}
\usepackage{subfigure}
\usepackage{tikz}
\usepackage{xcolor}

\usepackage{booktabs,caption,fixltx2e}
\usepackage[flushleft]{threeparttable}

\hypersetup{
 colorlinks=true,
 linkcolor=blue,
 citecolor=blue,
 filecolor=black,
 urlcolor=black,
 pdfauthor = {},
 pdftitle = {},
 pdfkeywords = {}
}

\usepackage{listings}
\lstdefinelanguage{json}{
    basicstyle=\ttfamily\small,
    numbers=left,
    numberstyle=\tiny,
    stepnumber=1,
    numbersep=5pt,
    showstringspaces=false,
    breaklines=true,
    stringstyle=\color{red},
    literate=
     *{0}{{{\color{blue}0}}}{1}
      {1}{{{\color{blue}1}}}{1}
      {2}{{{\color{blue}2}}}{1}
      {3}{{{\color{blue}3}}}{1}
      {4}{{{\color{blue}4}}}{1}
      {5}{{{\color{blue}5}}}{1}
      {6}{{{\color{blue}6}}}{1}
      {7}{{{\color{blue}7}}}{1}
      {8}{{{\color{blue}8}}}{1}
      {9}{{{\color{blue}9}}}{1}
      {:}{{{\color{black}{:}}}}{1}
      {,}{{{\color{black}{,}}}}{1}
      {\{}{{{\color{black}{\{}}}}{1}
      {\}}{{{\color{black}{\}}}}}{1}
      {[}{{{\color{black}{[}}}}{1}
      {]}{{{\color{black}{]}}}}{1},
}

\begin{document}

\title{Characterizing Large Language Model Agentic Workflows: A Study on N8n Ecosystem}

\titlerunning{Characterizing Large Language Model Agentic Workflows}        

\author{Yutian Tang       \and
        Yuming Zhou \and
        Huaming Chen 
}


\institute{Y. Tang \at
              School of Computing Science, University of Glasgow \\
              \email{Yutian.Tang@glasgow.ac.uk} \\
              Y.Tang is the corresponding author
           \and
           Y. Zhou \at %
           School of Computer Science, Nanjing University \\
           \email{zhouyuming@nju.edu.cn}
           \and 
           H. Chen \at
           The University of Sydney\\
           \email{huaming.chen@sydney.edu.au}
}

\date{Received: date / Accepted: date}

\maketitle

\noindent\textbf{Clinical trial number:} not applicable.

\begin{abstract}
Large Language Models (LLMs) are rapidly being adopted in low-code and no-code automation platforms, where non-expert users design workflows that combine natural language understanding with external services and APIs. LLM agents are LLM systems that use LLMs as a core "brain" to reason, plan, and autonomously execute complex, multi-step tasks. Although a large number of standardised evaluation benchmarks have emerged in recent years, these benchmarks mainly focus on assessing models' capabilities in single tasks such as knowledge comprehension, code generation, and mathematical reasoning; however, there remains a lack of systematic empirical research and large-scale analysis regarding how LLM agents are actually deployed and their operational characteristics within real-world workflow ecosystems. 

In this paper, we present the first large-scale empirical study of LLM agentic workflows in low-code automation platforms. We analyze more than 6,000 publicly available n8n workflows and examine four aspects of their design: task distribution, structural and tool use patterns, reliability mechanisms, and autonomy levels. Our analysis shows that LLM workflows are not merely prompt response pipelines. Instead, LLMs are commonly embedded within broader automation structures involving control logic, external tools, communication services, storage systems, and human review points. We further find that while many workflows include lightweight post-processing or routing logic after LLM execution, explicit reliability mechanisms such as structured fallback paths, repair loops, failure-specific alerts, and human approval gates remain relatively uncommon. These results reveal a gap between the increasing deployment of LLM agents in practical automation ecosystems and the limited engineering support for reliability, safety, and governance. Overall, our study provides ten empirical findings and five research takeaways for researchers, platform developers, and practitioners seeking to understand and improve real-world LLM agentic workflows.

\end{abstract}


\section{Introduction}

Large Language Models (LLMs) have rapidly advanced the state of the art in natural language processing. They are widely used in diverse tasks in software engineering, including code generation, test case generation, code summarization, and so on. Nowadays, LLMs are widely used as agents, where third-party tools or libraries are connected with language models to provide hands-on service to end users. For example, an LLM-powered coding assistant can act as an agent by invoking a Python interpreter to execute generated code, fetching live documentation from the web, or querying a database to retrieve real-time project context. Instead of merely focusing on a solo task, such as fixing a bug or generating a code snippet, an agent can autonomously run tests, analyze stack traces, and iteratively refine the code using external tools. However, the practical effectiveness of LLM agents in real‑world workflow automation environments (such as n8n) remains underexplored. This gap matters because an agent operating inside such an environment faces constraints rarely studied in isolation.

LLM agent workflows differ fundamentally from traditional single-task LLM applications. In a real-world agent workflow, an agent is not merely prompted to generate a textual answer or a code snippet. Instead, it is embedded in a workflow graph that contains multiple interconnected nodes, each representing a concrete operation, such as sending an HTTP request, querying a database, invoking a code execution module, applying a conditional branch, or waiting for an external trigger. In practice, these nodes are implemented as tools or APIs. The behaviour of the agent is therefore advanced not only by the underlying language model, but also by the workflow structure, node configuration, data dependencies, execution order, and runtime context. For example, when we ask a coding agent to build a web crawler for a website, it may first retrieve the structure of the target website, inspect relevant pages, decide on a framework to adopt, and then generate the code accordingly. It may further execute the generated code, observe runtime behaviour, and adjust or repair the code accordingly. Similar hybrid designs have been adopted in software repair, where LLM based reasoning is incorporated into a broader repair process rather than being treated as an isolated text generation step \cite{liu2026llm}. This indicates that the behaviour oaf an LLM agent is advanced and enhanced not only by the capability of the model, but also by the workflow structure, configurations, tools available, and runtime feedback.

\noindent\textbf{Motivation.} A systematic characterisation of LLM agent workflows is therefore needed. Such a characterisation should examine what kinds of nodes are commonly used, how tools are composed, how data flows across nodes, where human intervention is required, and what types of failures emerge during execution. For example, an agent may select the correct tool but pass an incorrect argument, retrieve relevant information but propagate it to the wrong downstream node, or generate a workflow that is syntactically valid but semantically inconsistent with the user’s intent. Studying and characterizing LLM agent workflows at the workflow level can reveal how agents actually operate in practical automation environments and provide empirical evidence for improving their reliability, usability, and safety.

\noindent\textbf{State-of-art.} Recent work has moved large language models from passive text generators toward agents that reason, plan, call tools, and interact with external environments. Toolformer trains models to decide when and how to call external tools such as calculators, search engines, translation systems, and calendars \cite{schick2023toolformer}. Prior work has made substantial progress on LLM reasoning, tool use, agent architectures, workflow generation, and benchmark-based evaluation \cite{yao2023react,schick2023toolformer,qin2024toolllm,wu2024autogen,fan2024workflowllm,liu2024agentbench,zhou2024webarena,jimenez2024swebench,xie2024osworld}. However, these studies provide limited empirical evidence about how LLM agents are actually assembled in public low-code workflow ecosystems.

\noindent\textbf{Research Questions.} To fill this gap, in this paper, we conduct a study to answer the following research questions: 

\noindent$\bullet$ \textbf{RQ1 (Tasks): What tasks are LLMs used for in real-world automation workflows?} In this RQ, we intend to understand task distribution of LLM workflow. It is important because it reveals the gap between benchmark-centric evaluation and practical adoption. It also helps identify the dominant application scenarios where future research on LLM agents, tooling, and evaluation should focus.

\noindent$\bullet$ \textbf{RQ2 (Structures \& Tools): What structural and tool-use patterns characterize LLM workflows?} In this RQ, we intend to characterize workflow structures and tool-use patterns. It can reveal common architectural designs, integration practices, and emerging agentic compositions adopted by practitioners.

\noindent$\bullet$ \textbf{RQ3 (Failure Handling): What reliability mechanisms are encoded around LLM outputs in workflow designs?} In this RQ, we intend to understand how workflow designers address reliability concerns.

\noindent$\bullet$ \textbf{RQ4 (Action Coupling and Human Control): To what extent are LLM outputs coupled to external actions, and where is human control placed in workflow designs?} In this RQ, we examine design level action coupling and human control placement. We do not claim to measure runtime autonomy. Instead, we analyse whether workflow JSON encodes paths from LLM nodes to external action nodes, and whether these paths pass through workflow logic or explicit human control mechanisms.

\noindent\textbf{Contribution.} In summary, in this paper, we make the following contributions:

\noindent$\bullet$ This is the \emph{first} large-scale empirical study of LLM-based automation workflows, analyzing task distributions, integration patterns, and failure-handling strategies to bridge the gap between benchmark evaluation and real-world deployment.

\noindent$\bullet $ We analyse 6,003 publicly available workflows on N8n to understand how LLMs workflows are designed.

\noindent$\bullet $ We investigate reliability and governance aspects of LLM workflows, including failure-handling mechanisms, autonomy patterns, and human oversight strategies, revealing how workflow authors balance automation, robustness, and control in practice.

\noindent$\bullet $ We release our dataset, analysis framework, and replication package to facilitate future research on workflow-aware LLM agents, automation systems, reliability engineering, and agent governance.

\section{Background}\label{sec:bg}

\noindent\textbf{LLMs in Workflow Automation.} Workflow automation systems are automated business processes that coordinate and control the flow of work and information between participant \cite{stohr2001workflow}. As Large Language Models (LLMs) are increasingly embedded into workflow automation systems, users are able to design their own pipelines that combine LLMs and other third-party services. In these low-code/no-code environments, LLMs are rarely used in isolation. Instead, they mainly serve as intermediate components that can transform unstructured text or data, translate and understand outputs given by services, or generate actions to trigger subsequent events. A typical workflow can chain multiple steps. For example, in a customer-support automation workflow, when a new support email arrives, the LLM agent can summarize the message, extract required actions, identify intent, and estimate urgency. Based on the sentiment, the workflow can then automatically flag high-priority events for users to follow up, and set up reminders.  

\begin{figure*}[!htpb]
    \centering
    \includegraphics[width=0.8\textwidth]{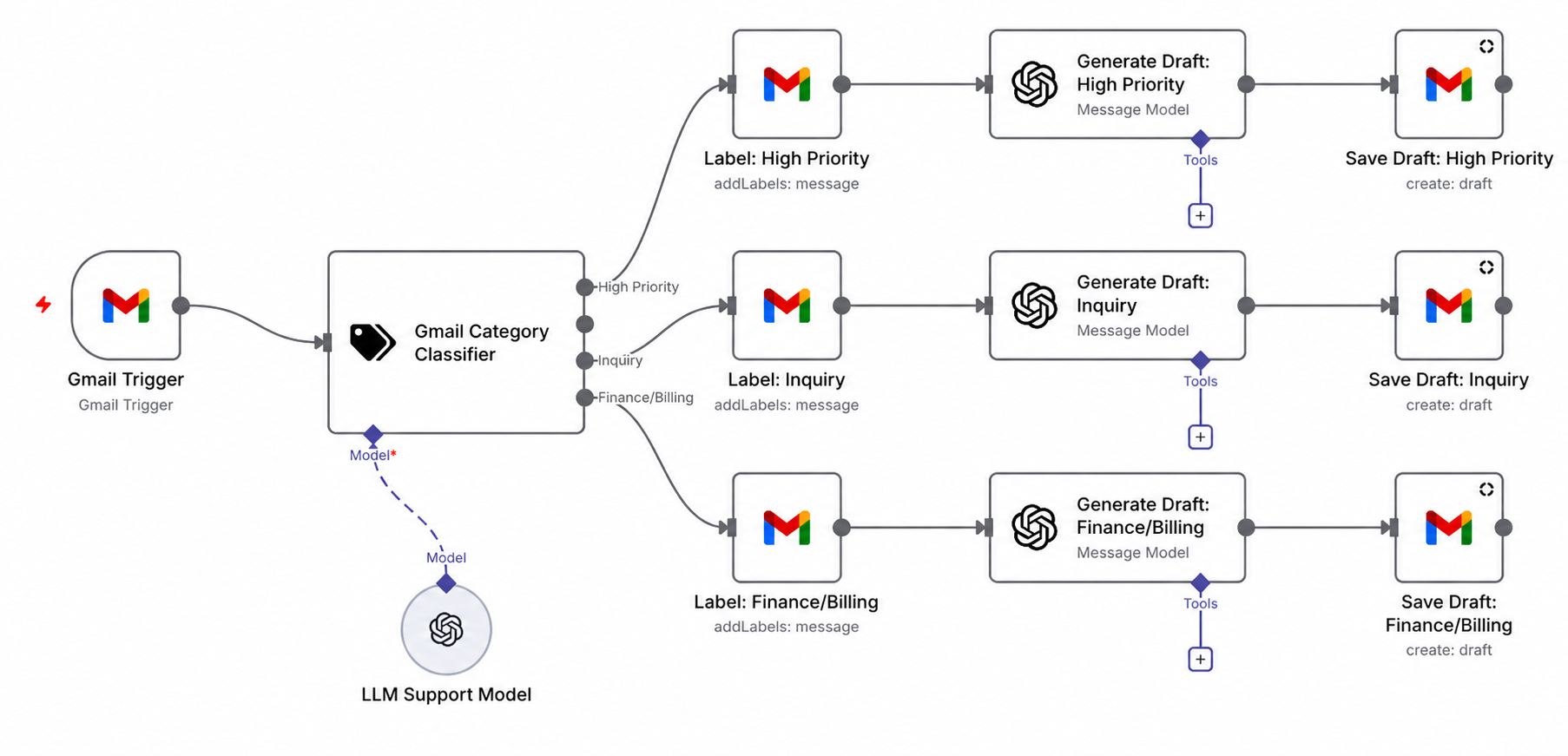}
    \caption{Gmail Email Classifier with GPT-4 Auto-Generated Draft Replies}
    \label{fig:gmailclassifier}
\end{figure*}

\noindent\textbf{Scope of this paper.} This paper focuses on LLM integration in workflow automation systems rather than on new LLM architectures or model level benchmarks. Our unit of analysis is the workflow design: how LLM nodes are connected to triggers, logic nodes, tools, external services, and reliability mechanisms. We therefore study publicly available n8n workflows as design artifacts that reveal how users compose LLMs with real automation infrastructure \cite{N8n}. N8n is an open-source workflow automation platform that enables users to connect applications, APIs, and services through a visual, node-based editor. Unlike proprietary tools such as Zapier, it can be self-hosted, giving organizations greater control over their data and integration environment. With a large and growing library of community-driven integrations, n8n provides a flexible and scalable way to design and automate workflows across diverse domains.

\noindent\textbf{Execution Model in N8n.} In N8n, workflows are represented as graphs of nodes and connections. In general, there are four node types\cite{N8n}: (1) \textit{app/action nodes}: app nodes perform operations within a workflow, such as, adding,  removing, and editing data, requesting and sending external data; and triggering events in other systems; (2) \textit{trigger nodes}: start a workflow and supply the initial data; (3) \textit{core nodes}: they can be either trigger or action nodes. Unlike most nodes, they aren't tied to a specific external service — instead they provide general functionality like logic, scheduling, or generic API calls (e.g., HTTP Request, If, Switch, Merge); and (4) \textit{AI cluster nodes}: groups of nodes that work together, primarily used in AI workflows. Each cluster has one root node and one or more sub-nodes that extend its functionality.

\noindent\textbf{Running Example of A Workflow.} Now, we will use  ``Gmail Email Classifier with GPT-4 Auto-Generated Draft Replies" \cite{N8n-8426} as an example from N8n to illustrate how a workflow works in practice. Fig. \ref{fig:gmailclassifier} shows the entire workflow. Specifically, incoming Gmail messages are polled every 15 minutes and passed to a lightweight LLM classifier that routes them into categories such as High Priority, Inquiry, or Finance/Billing. Note that some settings/parameters (e.g, every 15 minutes) can be retrieved from the corresponding JSON description file of the workflow (see. Listing \ref{lst:n8n8426}) rather than directly presented in the workflow image. 

The JSON representation of a workflow records the workflow graph, node types, node parameters, execution settings, and connection types. This makes it possible to statically inspect how LLMs are embedded into workflow designs. For example, JSON files reveal whether an LLM node is followed by an IF node, whether an AI Agent is connected to a tool, whether retry settings are enabled, or whether a structured output parser is attached. However, JSON does not fully reveal runtime behaviour, such as whether a fallback branch is actually triggered under realistic inputs. Therefore, our analysis treats JSON as evidence of encoded workflow design rather than as proof of runtime correctness. 

Each part invokes a tailored prompt: urgent emails generate concise action-oriented drafts with clarifying questions, inquiries receive polite and complete responses with next steps, and billing messages produce structured replies confirming amounts, due dates, and invoice IDs. Then, the system applies the relevant Gmail label and stores the reply as a draft in the original thread.

\begin{lstlisting}[language=json,caption={Gmail Auto-Generated Draft Replies Corresponding JSON Data},label={lst:n8n8426}]
{
...
  "pollTimes": {
    "item": [{ 
    "mode": "everyX", 
    "unit": "minutes", 
    "value": 15 } ]
    }
  ...
}
\end{lstlisting}

In this example, it demonstrates how LLMs can function as workflow agents rather than stand-alone chatbots: combining classification, prompt-conditioned generation, and draft persistence into a reproducible pipeline. It also highlights measurable aspects: classification accuracy, edit distance between drafts and final sends, or latency and cost per category to make such workflows useful as testbeds for evaluating LLMs in practice and real-world settings. 

\section{Empirical Study and Finding}
\subsection{Dataset}

\noindent$\bullet$ \textbf{Data Source.} Our study is based on publicly available n8n workflow templates that integrate Large Language Models. In n8n, a workflow is a collection of connected nodes that automate a process, and workflows can be exported and imported as JSON files. We therefore use workflow JSON files as the primary analysis artifact. The collected workflows come from the public n8n template ecosystem and represent reusable examples shared by users for automation, customization, and knowledge exchange.

\noindent$\bullet$ \textbf{Filtering Criteria.} Not all public n8n workflows use LLMs. To construct the dataset, we applied three filtering steps:

\begin{itemize}
    \item First, we selected workflows that contain at least one LLM-related component, including explicit LLM provider nodes, AI agent nodes, chat model nodes, or HTTP and API calls that match known text generation or chat completion endpoints. Examples include OpenAI, Anthropic, Hugging Face, Gemini, Mistral, Ollama, LangChain, AI Agent, and chat model components. This ensures that the selected workflow is an LLM-based workflow;

    \item Second, we excluded workflows without meaningful execution logic, such as workflows with isolated placeholder nodes, disconnected templates, or fewer than two executable nodes. We avoid filtering out simple but valid LLM workflows merely because they have few nodes, since workflows such as Webhook to LLM to Email or LLM to Slack can still represent real LLM automation usage; and

    \item Third, we removed duplicate workflows to reduce redundancy. Exact duplicates were identified using canonicalized JSON hashes, while near duplicates were identified using normalized workflow structure, node types, selected node parameters, and connection similarity.
\end{itemize}

\noindent$\bullet$ \textbf{Preprocessing and Normalization.} As we demonstrated in the running example in Sec. \ref{sec:bg}, each workflow is stored in JSON format, specifying nodes, parameters, and connections. We parsed these files into a normalized representation that captures: (1) the sequences of tasks; (2) the role of LLM components; and (3) the interactions among these components and external services.  In this representation, nodes capture workflow components, while connections capture execution and AI dependency relationships. We distinguish ordinary main execution connections from AI specific dependency connections, such as language model, tool, memory, output parser, embedding, document, retriever, and vector store links. This distinction is necessary because ordinary execution edges describe workflow control flow, whereas AI dependency edges describe how AI components are configured. We also remove non execution nodes, such as sticky notes and no operation nodes, from the main execution graph because they annotate workflows rather than execute behaviour.

\noindent$\bullet$ \textbf{Dataset Statistics and Quality Checks.} After filtering and preprocessing, our final dataset contains 6,003 valid workflow JSON files. These workflows contain 142,181 raw nodes, including 104,743 execution nodes and 37,438 non execution nodes. During graph construction, unresolved or ambiguous connections are recorded as diagnostics and excluded from graph construction. This prevents invalid references, removed annotation nodes, or duplicate display names from incorrectly affecting the workflow graph.

\subsection{RQ1: Tasks of LLMs}
\noindent \textbf{RQ: What tasks are LLMs used for in real-world automation workflows?} 

\noindent \textbf{Motivation.} Although numerous benchmarks have been proposed to evaluate LLM capabilities, most focus on isolated tasks such as question answering, code generation, summarization, or reasoning. Software engineering evaluations have similarly examined isolated tasks such as unit test generation, comparing LLM generated tests with established automated testing techniques in terms of correctness, readability, coverage, and defect detection \cite{tang2024chatgpt}. Understanding task distribution is important because it reveals the gap between benchmark-centric evaluation and practical adoption. It also helps identify the dominant application scenarios where future research on LLM agents, tooling, and evaluation should focus.

\noindent \textbf{Methodology.} 
To understand how LLMs are used in real-world settings and answer \textbf{RQ1}, we first established a taxonomy of task types. 

We adopt a task taxonomy grounded in standard NLP tasks  (e.g., classification, information extraction) \cite{JM3}, as well as the demonstrated capabilities of large language models in  text generation, translation, and question answering \cite{Brown:20}. We also include planning and agentic execution as a category as shown in prior work \cite{Yao:22}:

\noindent \textbf{(1) Text Generation:} Text generation takes text (e.g., a sequence, keywords) as input, processes the input into semantic representations, and generates desired output text \cite{Yu:2022}.

\noindent \textbf{(2) Question Answering:} Answering queries based on the given context or external knowledge base (e.g., RAG).

\noindent \textbf{(3) Information Extraction:} Extracting structured information from unstructured text.

\noindent \textbf{(4) Summarization:} Summarizing long content into a shorter representation.

\noindent \textbf{(5) Classification:} Assigning predefined labels to inputs.

\noindent \textbf{(6) Data Analysis and Reporting:} Analysing data and generating insights from the data.

\noindent \textbf{(7) Planning and Agentic Execution:} Performing multi-step reasoning, tool selection, and execution.

\noindent \textbf{(8) Translation:} Translating text across different languages.

\noindent \textbf{(9) Moderation and Safety:} Detecting harmful, unsafe, or policy-violating content.

To answer RQ1, we analyzed 6,003 LLM-related workflows extracted from the n8n template repository and classified the primary task performed by task-bearing AI components within each workflow. However, the raw workflow data (in JSON) is not annotated with tasks or categories. To annotate workflow tasks, we considered three annotation strategies. The first strategy is \textbf{keyword-based annotation}, where task labels were assigned using manually defined keyword rules over workflow names, node names, and prompt-related fields. The second strategy was zero-shot annotation, where a pretrained model was used to assign task labels without task-specific fine-tuning. In practice, we choose an NLI-based zero short classifier (i.e., \code{MoritzLaurer/deberta-v3-large-zeroshot-v2.0} model) for this approach. The third strategy was LLM-based annotation. For each workflow, we constructed a compact workflow summary from the JSON representation, including the workflow name, relevant AI nodes, prompt snippets, and surrounding node context. 

However, in practice, we find that the keyword-based method achieved only an accuracy of 40.5\% (81 out of 200) based on 200 randomly selected samples. We find that this method systematically over-assigned workflows to information extraction, because many workflow descriptions contain generic terms such as extract, parse, retrieve, or process, even when the primary task is text generation. Similarly, the zero-shot classifier only achieved an overall accuracy of 14.5\% (29 out of 200 workflows). Many workflows involving content creation, summarization, question answering, information extraction, and agent-based automation were incorrectly classified as code-related tasks. This suggests that the semantic representations used by the zero-shot classifier were unable to distinguish between software engineering workflows and general-purpose LLM applications reliably. These findings indicate that the zero-shot NLP approach is insufficient for accurately categorizing LLM-based automation workflows in the n8n ecosystem. 

As a result, we decided to use an LLM to assign the dominant primary task label. In practice, we leverage DeepSeek's latest model \code{deepseek-v4-flash} to annotate workflows.

\noindent \textbf{Results.} 

\begin{figure}[!htbp]
    \centering
    \includegraphics[width=\linewidth]{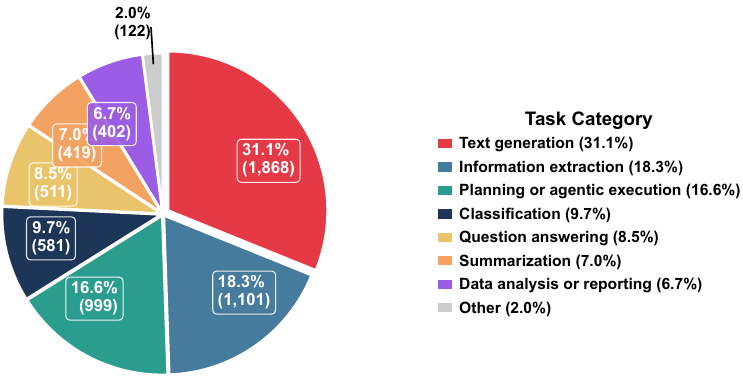}
    \caption{Primary LLM task distribution in n8n automation workflows.}
    \label{fig:rq1_task_distribution}
\end{figure}

As shown in Fig. \ref{fig:rq1_task_distribution}, LLMs are most frequently used for text generation, which accounts for 1,868 out of 6,003 workflows (31.12\%). This suggests that LLMs are frequently used as content production components, for example, to draft emails, generate reports, write social media posts, or produce marketing content. The second most common task is information extraction, covering 1,101 workflows (18.34\%), followed by planning or agentic execution with 999 workflows (16.64\%). Classification accounts for 581 workflows (9.68\%), while question answering accounts for 511 workflows (8.51\%). Summarization and data analysis or reporting account for 419 (6.98\%) and 402 (6.70\%) workflows, respectively. In contrast, code generation or transformation, translation, and moderation or safety are relatively rare, together accounting for less than 2.1\% of the workflows.

\noindent \textbf{Validation.} To evaluate the correctness of three approaches, we randomly select 200 workflows (detailed information can be found in our online artefact) to annotate them manually. 

The LLM-based annotation successfully labels 192 out of 200 workflows, resulting in an accuracy of 96\%. For the remaining eight workflows, the disagreements mainly arose from ambiguous or overlapping task boundaries, rather than clear annotation failures. For example, workflows that generate analytical reports may be reasonably interpreted as either text generation or data analysis, while assistant-style workflows may overlap between question answering and planning or agentic execution.

\begin{tcolorbox}[
  boxrule=1pt,
  boxsep=1pt,
  left=2pt,
  right=2pt,
  top=2pt,
  bottom=2pt,
  breakable,
  title=Answer to RQ1 (Tasks),
]

LLMs in n8n automation workflows are primarily used for text generation, information extraction, and planning or agentic execution. Among the 6,003 LLM-related workflows, text generation is the most frequent primary task, appearing in 1,868 workflows (31.12\%), followed by information extraction in 1,101 workflows (18.34\%) and planning or agentic execution in 999 workflows (16.64\%). This indicates that LLMs are mainly used as semantic processing and content generation components within automation pipelines, rather than only as chatbot-style question answering systems. Question answering accounts for 511 workflows (8.51\%), which shows that conversational use is important but not dominant.

\end{tcolorbox}

\subsection{RQ2: Structural and Tool-Use Patterns of LLM-Based Workflows}
\noindent \textbf{RQ: What structural and tool-use patterns characterize LLM workflows?} 

\noindent \textbf{Motivation.} LLM workflows are fundamentally different from standalone prompt-response applications because their behaviour is shaped not only by the language model itself, but also by workflow topology, external services, control logic, and tool integrations. Characterizing workflow structures and tool-use patterns can reveal common architectural designs, integration practices, and emerging agentic compositions adopted by practitioners.

\noindent \textbf{Methodology.} To characterize the topology and tool integration of real-world LLM workflows, we perform a static graph analysis over 6,003 valid n8n workflow JSON files (Table~\ref{tab:rq2_accounting}). We distinguish 104,743 execution nodes from 37,438 non-execution nodes. We exclude non-execution nodes from the main execution analysis because they annotate workflows rather than implement executable behaviour, while AI auxiliary components are retained for the AI dependency graph.

\begin{table}[!htbp]
\centering
\caption{Dataset and graph accounting for RQ2.}
\label{tab:rq2_accounting}
\begin{tabular}{lr}
\toprule
\textbf{Metric} & \textbf{Count} \\
\midrule
Workflow JSON files & 6,003 \\
Raw nodes & 142,181 \\
Execution nodes & 104,743 \\
Non-execution nodes & 37,438 \\
Main graph nodes & 83,201 \\
Main execution edges & 83,305 \\
AI auxiliary nodes & 21,228 \\
AI dependency edges & 20,728 \\
Agent composition instances & 8,849 \\
Workflows with LLM in main graph & 5,022 \\
\bottomrule
\end{tabular}
\end{table}

Although the dataset contains 6,003 LLM-related workflows, only 5,022 workflows contain an LLM node in the main execution graph. This is because our analysis separates ordinary execution flow from AI dependency configuration. Some workflows include LLM-related components only as AI auxiliary nodes, such as chat models, tools, and memory modules. These components make the workflow LLM-related, but they do not necessarily appear as nodes on ordinary main execution paths. Therefore, the 5,022 workflows represent cases where LLM participation is directly observable in the main execution layer, rather than the total number of LLM-related workflows in the dataset.

Two complementary graph views are constructed: (1) the \emph{main execution graph}, capturing ordinary control flow via main connections (83,201 nodes, 83,305 edges), and (2) the \emph{AI dependency graph}, modeling configuration links between AI root nodes and auxiliary components, such as \texttt{ai\_languageModel}, \texttt{ai\_tool}, and \texttt{ai\_memory}. 

Node analysis proceeds in two orthogonal steps. First, based on graph participation, each execution node is assigned to one of four \emph{analytical roles}: main-execution-only (78.56\%), AI-auxiliary-only (19.39\%), dual-role (0.88\%), or neither (1.18\%). Second, independently of role assignment, nodes are labeled with \emph{functional categories}, such as \texttt{trigger}, \texttt{llm\_ai}, and \texttt{logic\_control}, using deterministic rules based on their native type field.

\begin{table*}[!htbp]
\centering
\caption{Node taxonomy used for RQ2 structural and tool use analysis.}
\label{tab:node_taxonomy}
\small
\begin{tabular}{p{0.18\textwidth} p{0.36\textwidth} p{0.36\textwidth}}
\toprule
\textbf{Category} & \textbf{Definition} & \textbf{Example native node types or integrations} \\
\midrule

\texttt{trigger} &
Nodes that initiate workflow execution by receiving an event, polling a source, or starting a scheduled execution. &
Manual Trigger, Webhook, Schedule Trigger, Cron, Form Trigger, Chat Trigger, Email Trigger, Gmail Trigger, Telegram Trigger, MCP Trigger \\

\texttt{llm\_ai} &
Nodes that directly invoke, configure, or execute LLM based functionality, including model calls, AI agents, LLM chains, text classification, extraction, summarisation, and AI tool nodes. &
OpenAI, AI Agent, LLM Chain, Conversation Chain, Summarisation Chain, Retrieval QA Chain, Information Extractor, Sentiment Analysis, Text Classifier, Chat Model OpenAI, Chat Model Anthropic, Chat Model Gemini, Chat Model Mistral, Chat Model Ollama, Tool HTTP Request, Tool Code, Tool Calculator, MCP Client Tool, Structured Output Parser, Autofixing Output Parser, Guardrails, Model Selector \\

\texttt{logic\_control} &
Nodes that transform data, control branching, merge paths, implement conditional execution, or manipulate intermediate workflow state. &
IF, Switch, Merge, Filter, Code, Function, Function Item, Set, Edit Fields, Item Lists, Split In Batches, Split Out, Aggregate, Wait, Stop And Error \\

\texttt{web\_api} &
Nodes that interact with generic web resources or external APIs, but are not identified as LLM specific API calls. &
HTTP Request, GraphQL, RSS Feed Read \\

\texttt{communication} &
Nodes that send, receive, or route communication through email, chat, messaging, or notification services. &
Gmail, Email, Send Email, Slack, Discord, Telegram, Twilio, SendGrid, Microsoft Outlook, Mailchimp \\

\texttt{storage\_database} &
Nodes that read from, write to, or update tabular stores, databases, or lightweight application databases. &
Google Sheets, Airtable, Notion, Postgres, MySQL, MongoDB, Redis, Supabase, SQLite \\

\texttt{file\_document} &
Nodes that operate on files, documents, drives, spreadsheets, or extracted document content. &
Google Drive, Dropbox, OneDrive, Read Write File, Spreadsheet File, Extract From File, Convert To File \\

\texttt{crm\_project} &
Nodes that connect workflows to customer relationship management, issue tracking, or project management systems. &
HubSpot, Salesforce, Pipedrive, Jira, Linear, Trello, Asana \\

\texttt{other} &
Execution nodes that do not match the above categories. These nodes are retained in the graph but not interpreted as a specific functional group. &
Unmatched community nodes, uncommon integrations, or custom nodes \\

\bottomrule
\end{tabular}
\end{table*}

To make the node categorisation procedure explicit, Table~\ref{tab:node_taxonomy} summarises the taxonomy used in our analysis. The taxonomy is designed for structural analysis rather than semantic task classification. Therefore, categories are assigned primarily according to native n8n node types and integration names, rather than inferred from prompts or natural language descriptions. Exact node type matches are applied first. If no exact match is found, we apply keyword-based matching over node type and node name. HTTP Request nodes are treated as \texttt{llm\_ai} only when their parameters indicate calls to known LLM endpoints, such as chat completion, completion, embedding, OpenAI, Anthropic, Gemini, Mistral, Cohere, Hugging Face, or Ollama APIs. Otherwise, HTTP Request nodes are classified as \texttt{web\_api}. Nodes that do not match any rule are assigned to \texttt{other}. This conservative design avoids interpreting arbitrary HTTP calls or generic processing nodes as LLM components.

For AI dependency analysis, we further normalise AI related nodes into finer roles, including \texttt{ai\_agent}, \texttt{ai\_chain}, \texttt{chat\_model}, \texttt{tool}, \texttt{memory}, \texttt{output\_parser}, \texttt{embedding}, \texttt{vector\_store}, \texttt{document}, and \texttt{retriever}. This finer role taxonomy is used only for the AI dependency graph and agent composition analysis. In contrast, the main execution graph uses the coarser functional categories in Table~\ref{tab:node_taxonomy}, because its purpose is to characterise workflow topology and tool integration at the execution level.

Therefore, based on the nature of the nodes, we can then extract structural and tool-use patterns from workflows.

\noindent\textbf{Results.}

\noindent$\bullet$ \textbf{Workflow Topology.} LLM-based workflows exhibit substantial structural diversity. Only 23.27\% follow a strict linear chain, while the majority contain branching and merging (22.32\%), cyclic structures (18.92\%), or multi-component topologies (18.74\%). This indicates that conditional routing, iterative processing, and non-trivial control flow are prevalent.

\vspace{0.4em}
\noindent\fbox{%
\parbox{0.97\linewidth}{\textbf{Finding 1:} In practice, LLMs are used in a more complex topology, such as, conditional routing, iterative processing, and non-trivial, which are more prevalent than simple one-pass pipelines.}
}

\vspace{0.4em}

\noindent$\bullet$ \textbf{LLM Neighbourhood Patterns.} 

\begin{figure}[!htbp]
    \centering
    \includegraphics[width=\linewidth]{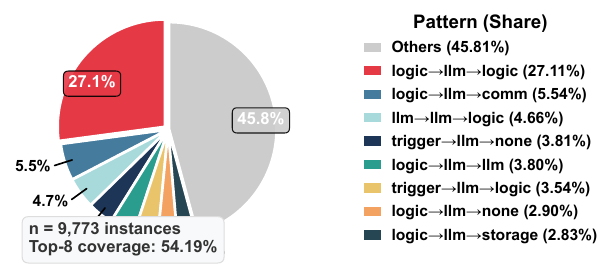}
    \caption{LLM Neighbourhood Patterns}
    \label{fig:rq2_llm_neighbourhood_patterns}
\end{figure}

To characterize how LLMs are contextually integrated, we analyse three-node sequences centred on \texttt{llm\_ai} nodes. The dominant pattern is \texttt{logic\_control} $\to$ \texttt{llm\_ai} $\to$ \texttt{logic\_control} (27.11\%), confirming that LLMs frequently serve as intermediate reasoning, classification, or extraction modules. As illustrated in Figure~\ref{fig:rq2_llm_neighbourhood_patterns}, the top eight neighbourhood patterns cover 54.19\% of all LLM contexts. Notably, patterns ending in \texttt{communication} (5.54\%) or \texttt{storage\_database} (2.83\%) demonstrate that LLM outputs are routinely consumed by downstream action or persistence components. For example, a customer-support workflow may filter tickets via an \texttt{IF} node, classify sentiment with an \texttt{OpenAI} node, and route results to escalation or auto-reply logic; alternatively, an \texttt{LLM Chain} may generate a summary that is directly sent to \texttt{Slack} or written to \texttt{Postgres}.

\vspace{0.4em}
\noindent\fbox{
\parbox{0.97\linewidth}{\textbf{Finding 2:} Neighbourhood pattern analysis reveals that LLMs are predominantly embedded as intermediate processing modules, with the dominant \texttt{logic\_control} $\to$ \texttt{llm\_ai} $\to$ \texttt{logic\_control} pattern accounting for 27.11\% of contexts. This indicates that LLM outputs are routinely consumed by downstream logic, communication, or storage components rather than serving as terminal endpoints.}
}

\vspace{0.4em}

\noindent$\bullet$ \textbf{AI Dependency Patterns.} 

\begin{figure}[!htbp]
    \centering
    \includegraphics[width=\linewidth]{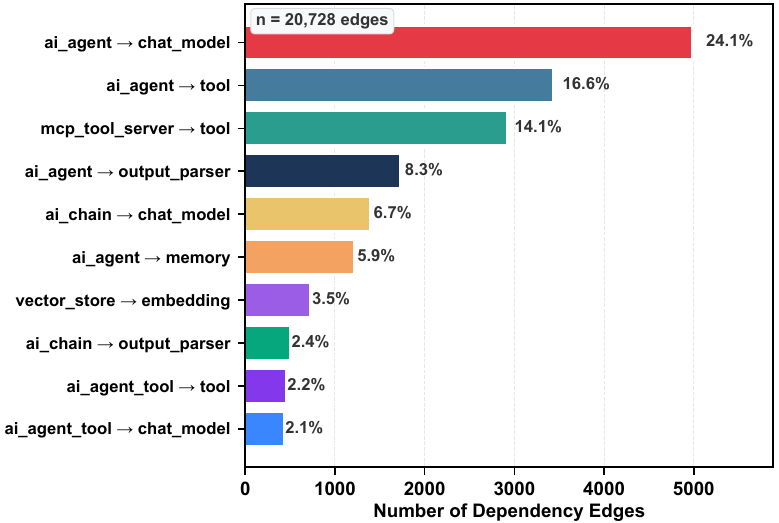}
    \caption{AI Dependency Patterns}
    \label{fig:rq2_ai_dependency_pie}
\end{figure}

The AI dependency graph reveals a second important aspect of workflow design: many workflows configure AI root nodes with auxiliary components rather than using isolated LLM calls. As shown in Fig.~\ref{fig:rq2_ai_dependency_pie}, the most frequent pattern is \texttt{ai\_agent} using \texttt{chat\_model} via \texttt{ai\_languageModel} (4,988 occurrences, 24.06\%), followed by tool integration via \texttt{ai\_tool} (3,435 for \texttt{ai\_agent}, 16.57\%; 2,916 for \texttt{mcp\_tool\_server}, 14.07\%). Other patterns include output parsing (8.34\%), memory modules (5.85\%), and retrieval-oriented embeddings (3.49\%). These results indicate that n8n workflows encode not only direct LLM API calls, but also tool-augmented, structured-output, stateful, and retrieval-oriented AI configurations at scale.

\vspace{0.4em}
\noindent\fbox{
\parbox{0.97\linewidth}{\textbf{Finding 3:} Real-world LLM workflows in n8n increasingly adopt \emph{agentic compositions} beyond direct model invocation. While basic \texttt{chat\_model} configurations dominate (24.06\%), a substantial share integrates auxiliary capabilities. This indicates a shift toward tool-augmented, structured, and stateful automation architectures at scale.}}

\vspace{0.4em}

\begin{figure}[!htbp]
    \centering
    \includegraphics[width=\linewidth]{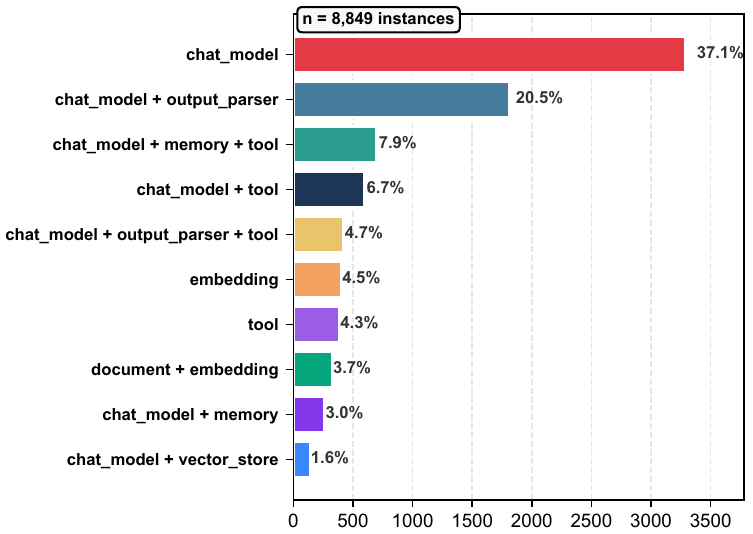}
    \caption{Agent Composition Patterns}
    \label{fig:rq2_agent_composition}
\end{figure}

\noindent$\bullet$ \textbf{Agent Composition.} To further understand how AI components are combined in practice, we analyse normalised agent composition patterns derived from the AI dependency graph. As shown in Fig. \ref{fig:rq2_agent_composition}, the most common configuration is a minimal setup with only a chat model, which appears 3,285 times and accounts for 37.12\% of all agent compositions.

However, a substantial number of workflows extend this basic configuration with structured output constraints. For instance, \textit{chat\_model + output\_parser} appears 1,812 times, representing 20.48\%. Tool augmented configurations are also present, although at a lower frequency. The pattern \textit{chat\_model + tool} appears 596 times, accounting for 6.74\%, while more complex configurations such as \textit{chat\_model + output\_parser + tool} appear 420 times, representing 4.75\%. 

More advanced agent designs, involving both memory and tools, are less common but still observable. For example, \textit{chat\_model + memory + tool} appears 696 times, accounting for 7.87\%.  In addition, non agent centric patterns such as \textit{embedding} (399 instances, 4.51\%) and \textit{document plus embedding} (323 instances, 3.65\%) show that some workflows focus on retrieval or document processing rather than conversational or agentic interaction.

\vspace{0.4em}
\noindent\fbox{\parbox{0.97\linewidth}{\textbf{Finding 4:} On one hand, many workflows use LLMs as standalone components. On the other hand, a smaller but meaningful subset adopts multi-component agent architectures combining models, tools, memory, and structured output. This distribution suggests that while advanced agentic workflows are present at scale, the dominant usage remains relatively lightweight.}}
\vspace{0.4em}

\begin{tcolorbox}[
  boxrule=1pt,
  boxsep=1pt,
  left=2pt,
  right=2pt,
  top=2pt,
  bottom=2pt,
  breakable,
  title=Answer to RQ2 (Structure \& Tools),
]

Real-world LLM workflows are structurally heterogeneous and functionally integrated. Rather than operating as simple prompt-to-response pipelines, they are predominantly logic-mediated automation systems where LLMs serve as intermediate processing nodes embedded within conditional and iterative control flows. Tool usage and agent composition have matured beyond basic model calls; a significant proportion of workflows incorporate output parsing, external tool calling, stateful memory, and retrieval components. These patterns are best captured through a dual-view analysis that jointly models the main execution topology and AI dependency configuration.
\end{tcolorbox}

\subsection{RQ3: Failure Handling}
\noindent \textbf{RQ: What reliability mechanisms are encoded around LLM outputs in workflow designs?} 

\noindent \textbf{Motivation.} LLM based workflows introduce reliability risks because an LLM node may execute successfully while still producing unreliable output, such as malformed JSON, missing fields or semantically incorrect content. This distinction between successful execution and output quality has also been observed in LLM based code generation, where generated artifacts must be evaluated separately in terms of correctness, complexity, and security \cite{liu2024no}. In workflow automation, such outputs are not isolated: they may be parsed, sent to users, or used to trigger external services. For example, an invoice workflow may write an incomplete amount field into a database, while a customer support workflow may route a wrongly classified request to the wrong communication channel. Therefore, RQ3 examines whether workflow designers encode reliability mechanisms around LLM outputs, such as structured output parsers, schema checks, post LLM parsing, conditional checks, fallback branches, human review, or repair loops. 

\noindent \textbf{Note.} Note that our analysis does not directly observe whether an LLM produces hallucinated content or whether such hallucinations are successfully detected at runtime. Hallucination is a semantic property of the generated output and usually requires comparison with external ground truth, task context, or human judgement. Workflow JSON can reveal whether designers encode mechanisms that may mitigate hallucination, such as retrieval steps, human review, or failure reporting. However, JSON alone cannot prove that these mechanisms identify factually incorrect or misleading LLM outputs. Therefore, RQ3 measures platform level reliability mechanisms and their structural association with LLM outputs rather than the actual effectiveness of hallucination detection or mitigation.

\noindent \textbf{Methodology.} 

Since workflow JSON captures node configurations and connections, our analysis identifies encoded reliability mechanisms rather than runtime reliability guarantees.

\noindent $\bullet$ \textbf{STEP 1:} We identify workflows that encode reliability-related mechanisms that are observable from workflow JSON. We identify platform-native mechanisms: \texttt{on\_error\_control}, \texttt{retry\_on\_fail}, \texttt{error\_trigger}, \texttt{stop\_and\_error}, \texttt{subworkflow\_error\_output}, and \texttt{always\_output\_data} \cite{N8n}. These anchors are detected from node-level settings, node types, and workflow-level error handling configuration. For example, a node with \texttt{retryOnFail} enabled, more than one \texttt{maxTries}, or a positive \texttt{waitBetweenTries} value is counted as \texttt{retry\_on\_fail}. A node with \texttt{onError}, \texttt{continueOnFail}, or error-output routing is counted as \texttt{on\_error\_control}. \texttt{Error Trigger} and \texttt{Stop And Error} nodes are detected from node types. Sub-workflow error outputs are detected when an execute-workflow-style node is configured with error-output or continue-on-failure behaviour. Finally, \texttt{always\_output\_data} captures nodes configured to continue producing output even when normal output data is missing. 

\noindent $\bullet$ \textbf{STEP 2:} In STEP 1, we identify workflows that contain platform-level error handling anchors, but we do not determine whether these anchors are related to LLM outputs. Therefore, we perform an error-anchor-centred backward traversal to identify error-handling anchors that are structurally associated with LLM outputs. For each workflow identified, we first construct the main execution graph using ordinary \texttt{main} connections. We then locate all error handling anchors identified in Step 1. Starting from each anchor, we traverse upstream along the reversed main execution graph. Instead of using a fixed hop limit as the primary stopping rule, we stop when the search first reaches one of three boundary nodes: an LLM node, a tool node, or an error handling node. If the first boundary reached is an LLM node, we keep the path and interpret the anchor as an LLM-associated error-handling anchor. If the first boundary is a tool node, we discard the path because the error handler is more likely to be associated with the tool failure rather than the LLM output. If the first boundary is another error handling node, we also discard the path to avoid chaining multiple error handlers into a single pattern.

\noindent \textbf{Results} Among the 6,003 filtered workflows, 1,829 workflows contain at least one platform-level error handling mechanism. Table \ref{tab:rq3-platform-anchors} shows the distribution of platform-level error handling anchors.

\vspace{0.4em}
\noindent\fbox{
\parbox{0.97\linewidth}{\textbf{Finding 5:} This distribution suggests that workflow authors more often rely on local continuation, retry, or output-preservation settings than on dedicated error workflow structures.}}
\vspace{0.4em}

\begin{table}[t]
\centering
\caption{Platform-level error handling anchors.}
\label{tab:rq3-platform-anchors}
\begin{tabular}{lrr}
\toprule
Anchor type & Count & Percentage among anchors \\
\midrule
\texttt{on\_error\_control} & 2,608 & 48.45\% \\
\texttt{retry\_on\_fail} & 1,854 & 34.44\% \\
\texttt{always\_output\_data} & 1,575 & 29.26\% \\
\texttt{stop\_and\_error} & 111 & 2.06\% \\
\texttt{error\_trigger} & 100 & 1.86\% \\
\texttt{subworkflow\_error\_output} & 31 & 0.58\% \\
\bottomrule
\end{tabular}
\end{table}

\begin{table*}[t]
\centering
\caption{Grouped LLM-associated error handling patterns.}
\label{tab:rq3-grouped-patterns}
\begin{tabular}{lrrrr}
\toprule
Pattern group & Paths & Ratio & Workflows & Anchors \\
\midrule
Direct LLM anchored error handling & 516 & 37.83\% & 359 & 490 \\
Parser-mediated LLM error handling & 238 & 17.45\% & 160 & 211 \\
Logic-mediated LLM error handling & 185 & 13.56\% & 123 & 169 \\
Parser-and-logic-mediated LLM error handling & 323 & 23.68\% & 166 & 236 \\
Long internal preprocessing before error handling & 102 & 7.48\% & 29 & 45 \\
\midrule
Total & 1,364 & 100.00\% & 684 & 1,123 \\
\bottomrule
\end{tabular}
\end{table*}

Out of 5,383 error handling anchors, 1,123 anchors are structurally associated with an upstream LLM. These anchors occur in 684 workflows and produce 1,364 LLM-associated error handling paths.

\vspace{0.4em}
\noindent\fbox{
\parbox{0.97\linewidth}{\textbf{Finding 6:} Among the 1,829 workflows with platform-level error handling, only 684 workflows contain error handling that can be conservatively linked to LLM outputs. This shows that only a subset are positioned close enough to LLM outputs to plausibly handle LLM-related failures.}
}
\vspace{0.4em}

\begin{table}[t]
\centering
\caption{LLM-associated error handling identified from platform-level anchors.}
\label{tab:rq3-llm-associated-error-handling}
\begin{tabular}{lr}
\toprule
Metric & Count \\
\midrule
Workflows with platform-level error handling & 1,829 \\
Error handling anchors & 5,383 \\
Anchors structurally associated with LLM outputs & 1,123 \\
Workflows with LLM-associated error handling & 684 \\
LLM-associated error handling paths & 1,364 \\
\bottomrule
\end{tabular}
\end{table}

We further group the 1,364 LLM-associated paths into five high-level patterns. As shown in Table \ref{tab:rq3-grouped-patterns}, the largest group is direct LLM anchored error handling, with 516 paths, accounting for 37.83\% of all LLM-associated paths. These paths appear in 359 workflows and 490 anchors. The second largest group is parser-and-logic-mediated error handling, with 323 paths, or 23.68\%. Parser-mediated handling accounts for 238 paths, or 17.45\%, while logic-mediated handling accounts for 185 paths, or 13.56\%. Finally, 102 paths, or 7.48\%, involve longer internal preprocessing chains before reaching error handling.

\vspace{0.4em}
\noindent\fbox{
\parbox{0.97\linewidth}{\textbf{Finding 7:} This indicates that LLM reliability handling is frequently embedded in local output-processing pipelines rather than being handled immediately at the LLM node. For example, parser-mediated paths suggest that workflow authors often treat unexpected LLM output as a parsing or transformation problem. Logic-mediated paths suggest that workflows sometimes encode conditional checks around LLM outputs before invoking platform-level error handling.}
}
\vspace{0.4em}

Although 684 workflows contain LLM-associated error handling, the structure is usually lightweight. The dominant anchors are still local settings such as \texttt{on\_error\_control}, \texttt{retry\_on\_fail}, and \texttt{always\_output\_data}. Stronger explicit mechanisms such as \texttt{stop\_and\_error}, \texttt{error\_trigger}, and \texttt{subworkflow\_error\_output} appear much less frequently.

\vspace{0.4em}
\noindent\fbox{
\parbox{0.97\linewidth}{\textbf{Finding 8:} Workflow authors are aware of failure handling, but often encode it through local node configuration rather than explicit end-to-end error recovery design.}
}
\vspace{0.4em}

\begin{tcolorbox}[
  boxrule=1pt,
  boxsep=1pt,
  left=2pt,
  right=2pt,
  top=2pt,
  bottom=2pt,
  breakable,
  title=Answer to RQ3 (Failure Handling),
]
Platform-level reliability mechanisms are present in a substantial minority of LLM workflows. Among 6,003 workflows, 1,829 contain platform-level error handling, while 684 contain error handling anchors that can be conservatively traced to LLM outputs. The most common LLM-associated pattern is direct error handling after an LLM, but most paths involve some intermediate parser, transformation, or logic node before error handling. This suggests that real-world n8n workflows often handle LLM reliability indirectly through local output processing and node-level error settings, rather than through systematic, explicit LLM failure recovery mechanisms.

\end{tcolorbox}

\subsection{RQ4: Action Coupling and Human Control} 
\noindent \textbf{RQ: To what extent are LLM workflows designed as autonomous versus human controlled systems?} 

\noindent \textbf{Motivation.} As LLMs increasingly participate in decision-making and action execution, an important governance question is whether their outputs remain internal to the workflow or can influence downstream external actions. In this RQ, we examine design level action coupling and human control placement. We do not claim to measure runtime autonomy. Instead, we analyse whether workflow JSON encodes paths from LLM nodes to external action nodes, and whether these paths pass through workflow logic or explicit human control mechanisms.

\noindent \textbf{Methodology.} To understand to what extent LLM workflows are designed as autonomous versus human-controlled systems, we first parse each workflow and extract its nodes and connections. Then, we annotate each node with one of several functional roles relevant to autonomy and human control. A node is marked as an LLM node if its type, name, or configuration indicates an LLM-related component. A node is marked as an external action node if it may interact with an external service or produce side effects, such as sending messages or sending emails. A node is marked as a control or validation node if it represents workflow logic, parsing, validation, or branching, such as \texttt{IF}, \texttt{Switch}, output parser, or guardrail nodes. For human-control nodes, we do not treat generic terms such as ``review'', ``manual'', or ``confirm'' as sufficient evidence of human control, because these terms may describe task content rather than actual human oversight. Instead, we require stronger indicators, such as human-in-the-loop tool nodes, approval-related node types, or explicit phrases such as ``human in the loop'', ``wait for approval'', ``manual approval'', ``approval required'', or ``review and approve''.

Next, we trace downstream execution paths starting from each LLM node. For each workflow, we build a directed execution graph and perform breadth-first search from every LLM node to identify whether its output can reach an external action node. In our implementation, the downstream search depth is capped at six hops to capture execution dependencies while avoiding unrelated distant paths in large workflows. Then, we record whether the path is direct or mediated by intermediate nodes:

\noindent $\bullet$ \textbf{Direct}: A path is a direct automated action when an LLM node connects directly to an external action node, for example \texttt{LLM} $\rightarrow$ \texttt{Slack}.

\noindent $\bullet$ \textbf{Workflow-logic Mediated}: A path is workflow-logic mediated when at least one intermediate control or validation node appears before the action, for example \texttt{LLM} $\rightarrow$ \texttt{IF} $\rightarrow$ \texttt{Gmail}.

\noindent $\bullet$ \textbf{Human Mediated}: A path is human mediated when at least one intermediate human-control node appears before the action, for example \texttt{LLM} $\rightarrow$ \texttt{Approval} $\rightarrow$ \texttt{Send Email}.

Finally, we aggregate the path-level categories into a workflow-level autonomy mode. Since a workflow may contain multiple LLM nodes and multiple LLM-to-action paths, we assign one primary label using a priority-based rule. If any path is human-mediated, the workflow is classified as \texttt{human\_gated}. Otherwise, if any path is workflow-logic mediated, the workflow is classified as \texttt{logic\_gated\_automation}. Otherwise, if any is direct, the workflow is classified as \texttt{automated\_action}. If the workflow contains LLM nodes but no downstream path from an LLM node to an external action node, it is classified as \texttt{low\_autonomy}. Workflows without detected LLM nodes are classified as \texttt{unresolved\_llm\_path}.

\noindent \textbf{Results.} 

\begin{table}[!htbp]
\centering
\caption{Workflow-level autonomy modes in LLM workflows.}
\label{tab:rq4-autonomy-modes}
\begin{tabular}{lrr}
\toprule
Autonomy mode & Workflows & Percentage \\
\midrule
\texttt{logic\_gated\_automation} & 2,509 & 41.80\% \\
\texttt{automated\_action} & 1,592 & 26.52\% \\
\texttt{low\_autonomy} & 1,363 & 22.71\% \\
\texttt{human\_gated} & 167 & 2.78\% \\
\texttt{unresolved\_llm\_path} & 372 & 6.20\% \\
\bottomrule
\end{tabular}
\end{table}

Table~\ref{tab:rq4-autonomy-modes} shows the workflow-level autonomy modes identified by our analysis. Out of 6,003 workflows, 2,509 workflows (41.80\%) are classified as \texttt{logic\_gated\_automation}, meaning that LLM outputs can reach external action nodes through workflow logic, parsing, validation, or control nodes. Another 1,592 workflows (26.52\%) are classified as \texttt{automated\_action}, where LLM outputs can reach external actions without detected human control or intermediate workflow control. In contrast, only 167 workflows (2.78\%) are classified as \texttt{human\_gated}. This indicates that explicit human control before external action is rare in the studied LLM workflows. 

\vspace{0.4em}
\noindent\fbox{
\parbox{0.97\linewidth}{\textbf{Finding 9:} Only 167 workflows (2.78\%) contain a detected human-mediated path before an external action. This suggests that explicit human approval or human-in-the-loop gating is not a common design choice in current n8n LLM workflows.}
}
\vspace{0.4em}

\vspace{0.4em}
\noindent\fbox{
\parbox{0.97\linewidth}{\textbf{Finding 10:}  The two largest categories are \texttt{logic\_gated\_automation} and \texttt{automated\_action}, covering 4,101 workflows (68.32\%). This shows that LLM outputs frequently participate in execution paths that can reach external services, such as email, messaging, HTTP APIs, storage, or database nodes.}
}
\vspace{0.4em}

\texttt{low\_autonomy} accounts for 1,363 workflows (22.71\%), where LLM nodes are present but no downstream path from an LLM node to an external action node is detected. Finally, 372 workflows (6.20\%) are labelled as \texttt{unresolved\_llm\_path}, because they were included in the LLM-based dataset but did not contain a traceable LLM execution node under our RQ4 analysis.

\noindent \textbf{Validation.} We manually select 200 sample to evaluate the accuracy of our tool. Specifically, for each workflow, we manually inspected downstream execution paths after LLM nodes, including external actions, control flow nodes, and human approval points, to determine whether the predicted autonomy mode is correct. The manual audit confirmed that the implemented detector correctly captured the analysed path structure in the sampled workflows. 

\begin{tcolorbox}[
  boxrule=1pt,
  boxsep=1pt,
  left=2pt,
  right=2pt,
  top=2pt,
  bottom=2pt,
  breakable,
  title=Answer to RQ4 (Action Coupling and Human Control),
]
LLM workflows are predominantly designed with action coupling rather than explicit human mediation. Human mediated action coupling appears in only 2.78\% of workflows, while logic mediated and direct action coupling together account for 68.32\%. This suggests that LLM outputs frequently reach downstream external services through automated workflow structures. However, these results should be interpreted as design level action coupling, not as evidence of runtime autonomy. Workflow JSON can reveal how LLM outputs are connected to external actions and human control points, but it cannot prove whether the LLM makes independent decisions under real inputs.

\end{tcolorbox}

\section{Case Study}
In this section, we conduct two case studies to future understand LLMs are used in real-world workflows.

\subsection{Case Study 1: Production AI Playbook: Human Oversight}
In this case study, we intend to study the workflow Production AI Playbook: Human Oversight (\#13849) \cite{N8n-13849} to illustrate a compact human gated design for LLM generated content.

\begin{figure}[!htbp]
    \centering
    \includegraphics[width=\linewidth]{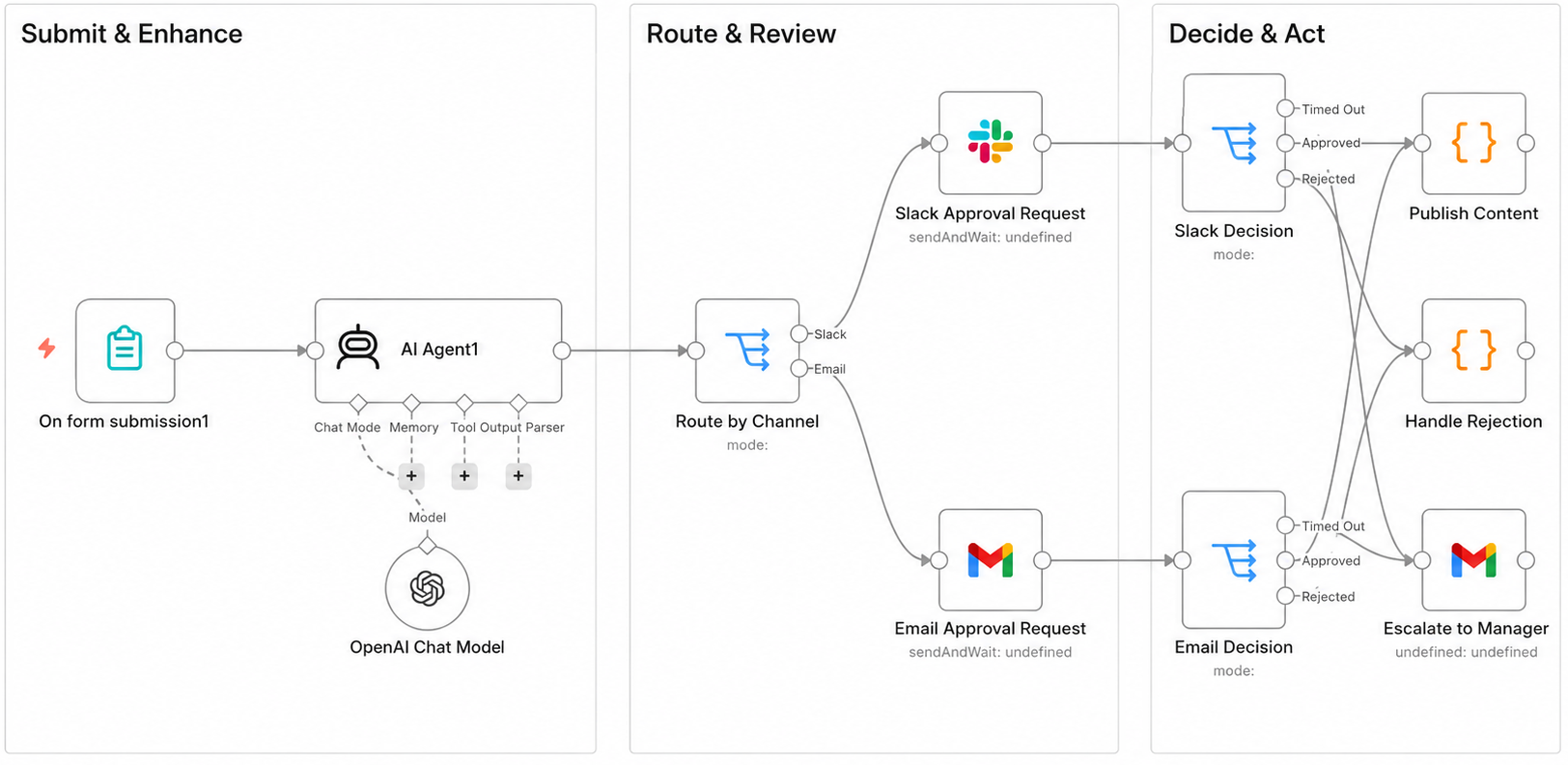}
    \caption{13849 Workflow}
    \label{fig:casestudy1}
\end{figure}

As shown in Fig.\ref{fig:casestudy1}, in the first stage (\textbf{Submit \& Enhance}), a submitted form is first given to LLM, which is connected to a ChatGPT model. In the second stage (\textbf{Route \& Review}), the AI Agent output first reaches \code{Route by Channel}, which dispatches the approval request through either Slack or Email. Then, it goes to the last stage (\textbf{Decide \& Act}), the Slack or Email approval leads to three possible outcomes: \textit{approved}, \textit{rejected}, and \textit{timed out}. The \textit{approved} path leads to publish content, the \textit{rejected} path leads to handle rejection, and \textit{timed out} leads to escalate to manager. 

\vspace{0.4em}
\noindent\fbox{
\parbox{0.97\linewidth}{\textbf{Takeaway 1: LLM output should not be treated as directly executable content.} The LLM output should be treated as a draft or proposal rather than a final action.}
}
\vspace{0.2em}

\vspace{0.4em}
\noindent\fbox{
\parbox{0.97\linewidth}{\textbf{Takeaway 2: External actions should be gated by human decisions.} Before invoking any external actions, an approval design should be considered. A strong approval design should not only consider approval and rejection, but other cases, such as, timeout, lack of response.}
}
\vspace{0.2em}

\subsection{Case Study 2: Invoice Payment Tracking with LLM Based Extraction and Post Processing} 

In this case study, we leverage an automated invoice payment tracking with OCR, Claude AI, Slack and Notion DB (\#7773) \cite{N8n-7773} to discuss how LLM output checks, downstream routing, and human review interact in a realistic document processing pipeline.

This workflow is triggered by invoices files uploaded through Slack. It first checks if the input is an image or a document, applies OCR, then sends the text to a Claude-based LLM chain for invoice field extraction. 

\begin{figure}[!htbp]
    \centering
    \includegraphics[width=\linewidth]{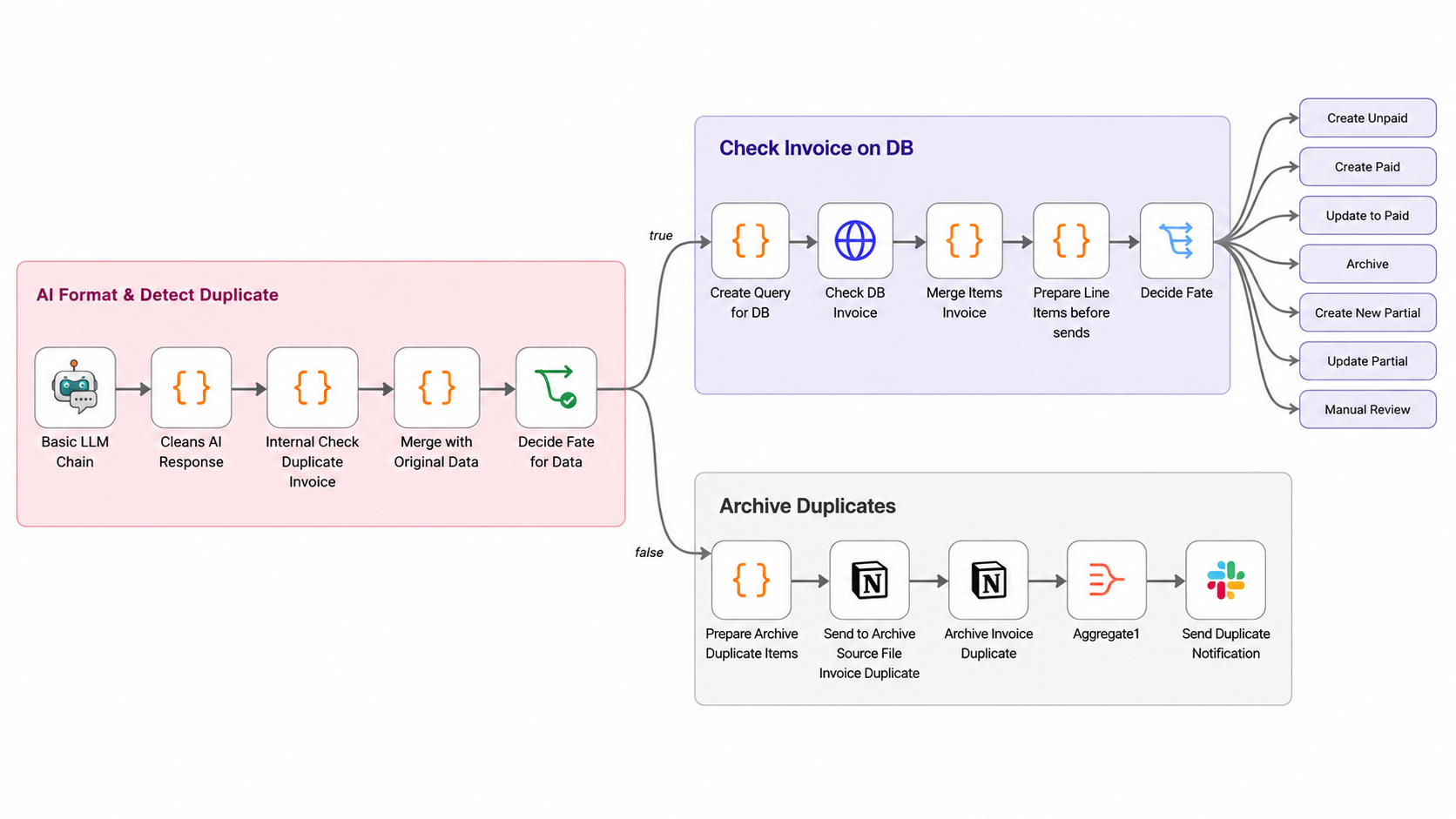}
    \caption{Core Component of 7773 Workflow}
    \label{fig:casestudy2}
\end{figure}

As the core component of this work is displayed in Fig. \ref{fig:casestudy2}, the LLM output is then processed by a dedicated \code{Cleans AI Response} node, which removes code fences, searches for the top level JSON block, parses JSON, normalises numeric fields, and so on. The workflow then performs two forms of containment before external propagation. First, it checks for internal duplicate invoices. Duplicate items are marked, and items labelled as drop are routed to an archive path. Second, for non duplicate items, the workflow builds a Notion database query and compares the extracted invoice against existing database records. This comparison produces a next\_action field, which is then consumed by a Decide Fate switch node.

This case is especially useful for RQ4 because it illustrates a mixed autonomy design. The LLM has substantial influence over the workflow because extracted fields such as amount\_total, amount\_due, and vendor affect downstream routing and database updates. However, the LLM does not directly perform external actions. Its output is mediated by code based repair, duplicate detection, Notion lookup, and a switch node.

\vspace{0.4em}
\noindent\fbox{
\parbox{0.97\linewidth}{\textbf{Takeaway 3: LLM input is a mediated representation.} LLM inputs in automation workflows should be understood as task specific representations produced by upstream preprocessing, rather than as the original user or system inputs. In this case study, the LLM does not operate directly on the uploaded invoice file. Its effective input is an OCR derived textual representation of the invoice. }
}
\vspace{0.2em}

\vspace{0.4em}
\noindent\fbox{
\parbox{0.97\linewidth}{\textbf{Takeaway 4: LLM output requires workflow level mediation before external execution} From the reliability perspective, outputs from LLMs are should not directly be used by downstream applications, instead, a series of workflow logics is applied to ensure correctness of the outputs before using them. }
}
\vspace{0.2em}

The workflow provides evidence of containment oriented post LLM checks, including JSON repair, schema normalization, duplicate detection, database matching, and conditional manual review. However, these checks mainly target malformed, inconsistent, duplicate, or ambiguous outputs. They do not constitute full semantic validation against the original invoice content. Therefore, in ambiguous cases, the workflow sends a Slack review message and waits for a human decision. The reviewer can approve the invoice as paid, approve it as unpaid, or archive it.

\vspace{0.4em}
\noindent\fbox{
\parbox{0.97\linewidth}{\textbf{Takeaway 5: Post LLM checks may not ensure semantic correctness.} In workflows, downstream checks can improve containment. However, these checks do not necessarily validate whether the LLM output is semantically correct with respect to the data source. In this case, human inference or LLM-as-a-judge is needed. }
}
\vspace{0.2em}
\section{Discussion}
\subsection{Threats to Construct Validity}
The main construct validity threat is whether static workflow JSON can accurately capture the concepts studied in this paper. Workflow JSON records node types, settings, parameters, and connections, which makes it suitable for analysing workflow design, graph structure, tool integration, and explicitly encoded reliability or autonomy mechanisms. However, it does not reveal runtime behaviour, actual LLM outputs, external service responses, or user intent. Therefore, our findings should be interpreted as design level evidence rather than direct measurements of production behaviour.

\subsection{Threats to Internal Validity}

The main internal validity threat concerns possible errors in our data processing and static analysis pipeline, including JSON parsing, node classification, graph construction, and mechanism detection. For example, an LLM call may be implemented through a generic HTTP node rather than an explicit LLM provider node, or a control node may be misclassified if its parameters are incomplete or ambiguous. Such errors could affect the measured frequency of workflow structures, tool use patterns, reliability mechanisms, and autonomy levels.

To reduce this threat, we use deterministic rules based on node types, settings, parameters, and connection context, and we separate main execution connections from AI dependency connections. This prevents AI configuration edges, such as a chat model connected to an agent, from being incorrectly treated as ordinary execution flow. We also use stable node identifiers to avoid merging nodes with duplicate display names, and we report unresolved or ambiguous connections as diagnostics rather than silently including them in graph construction. Nevertheless, some custom nodes, incomplete templates, or unconventional workflow designs may still be imperfectly classified.

\subsection{Threats to External Validity}

Our dataset represents only workflows made publicly available by users. Many production workflows remain private, thus our dataset may not fully represent the complete usage landscape. Public templates may also be biased toward examples, tutorials, demonstrations, or reusable templates. In addition, workflow JSON files do not contain all runtime information, such as private credentials, live execution traces, external service states, or actual LLM outputs. Therefore, our findings should be interpreted as evidence of design patterns and practices observable in the public workflow ecosystem rather than as a comprehensive characterization of the entire workflow automation landscape.

\section{Related Work}

\subsection{LLM Agents and Tool Use}

Recent work has moved large language models from passive text generators toward agents that reason, plan, call tools, and interact with external environments. Toolformer trains models to decide when and how to call external tools such as calculators, search engines, translation systems, and calendars \cite{schick2023toolformer}. Gorilla focuses on API calling and shows that retrieval-augmented API documentation can reduce hallucinated API calls \cite{patil2023gorilla}. API Bank, ToolLLM, and ToolBench provide benchmarks and datasets for evaluating tool selection, API retrieval, and tool invocation across large API spaces \cite{li2023apibank,qin2024toolllm,qin2023toolbench}. HuggingGPT connects LLMs with specialized models from Hugging Face and treats the LLM as a controller for task planning and model selection \cite{shen2023hugginggpt}. MRKL systems similarly combine neural models with symbolic tools and external modules \cite{karpas2022mrkl}. Hybrid architectures have also emerged in software security, where LLM assistance is combined with inter procedural, path sensitive taint analysis rather than replacing deterministic program analysis\cite{ji2025artemis}. These studies show that external tools can substantially extend LLM capabilities. At the same time, tool augmentation introduces security risks that cannot always be observed from static workflow structure, including runtime exposure of external inputs, local files, and environment data to LLM relevant outputs \cite{tan2026mcp}. However, their primary concern is whether models can select or call tools correctly. 

Several recent systems also study agent execution and self-improvement. Reflexion lets agents improve through verbal feedback and memory \cite{shinn2023reflexion}. Self-Refine iteratively improves model outputs using model-generated feedback \cite{madaan2023selfrefine}. Voyager demonstrates continual skill acquisition in an open-ended Minecraft environment \cite{wang2024voyager}. Generative Agents introduce memory, reflection, and planning for believable social simulation \cite{park2023generative}. AutoGen proposes a framework for composing multiple conversable agents that may use LLMs, tools, code execution, and human input \cite{wu2024autogen}. These systems focus on agent architecture and runtime interaction. Our work is complementary because we do not introduce a new agent framework. Instead, we conduct a large-scale empirical study of how existing LLM-based agents and workflows are designed in a low-code automation ecosystem.

\subsection{Workflow Orchestration and Low-Code Automation}

Recent empirical research has distinguished traditional low code development from LLM based low code development, identifying differences in their usage scope, limitations, and integration with visual programming and LLM agents \cite{liu2025empirical}. Workflow-based automation differs from standalone tool calling because the behavior of the system is determined not only by the model but also by workflow topology, node configuration, data dependencies, and execution order. 

WorkflowLLM is closest to our setting. It studies LLM workflow orchestration and constructs WorkflowBench from Apple Shortcuts and RoutineHub workflows \cite{fan2024workflowllm}. Its goal is to improve LLMs' ability to generate workflow orchestration plans. By contrast, our paper does not train an LLM to generate workflows. We study public n8n workflows as empirical artifacts and ask what task types, graph structures, tool integrations, reliability mechanisms, and autonomy patterns appear in the wild. This distinction is important. 

\subsection{Benchmarks for Realistic Agent Behavior}

A large body of work evaluates LLM agents in controlled environments. AgentBench evaluates LLMs as agents across multiple environments and highlights challenges in long-horizon reasoning and decision making \cite{liu2024agentbench}. WebArena provides realistic web environments for evaluating autonomous web agents \cite{zhou2024webarena}. MiniWoB and WebShop evaluate web interaction and shopping tasks \cite{shi2017world,yao2022webshop}. ALFWorld and ScienceWorld evaluate embodied or scientific interactive tasks \cite{shridhar2021alfworld,wang2022scienceworld}. SWE-bench evaluates whether models can solve real GitHub issues, while SWE-agent studies agentic software engineering workflows built around repository interaction, tool execution, and iterative repair \cite{jimenez2024swebench,yang2024sweagent}. OSWorld extends agent evaluation to real computer environments and shows that open-ended computer control remains difficult for current agents \cite{xie2024osworld}.

These benchmarks are essential because they show that agent performance drops sharply when tasks require long-horizon interaction, state tracking, external tools, and environmental feedback \cite{liu2024agentbench,zhou2024webarena,jimenez2024swebench,xie2024osworld}. However, benchmark studies usually define tasks and environments from the researcher side. Our work instead studies naturally occurring workflow artifacts from a public automation ecosystem. This gives a different kind of evidence. Rather than asking whether an LLM agent can complete a task in a benchmark, we ask how real workflows are assembled, which external services are connected, where LLMs sit in the execution graph, and whether workflow authors encode reliability or human oversight mechanisms.

\subsection{Human Control, Autonomy, and Governance}

Human control is another key issue in LLM-based automation. In software engineering, human in the loop systems have also incorporated feedback from software quality assurance personnel into online prediction processes, demonstrating that human intervention may improve both system performance and operational decision making \cite{liu2025human}. Multi-agent and tool-using frameworks often allow human input as part of the agent loop \cite{wu2024autogen}, and human feedback has been widely studied for aligning or improving model behavior \cite{shinn2023reflexion,bai2022constitutional,ouyang2022training}. In deployed automation systems, however, human control can appear in several structurally different forms. A workflow may be fully automated, where an LLM output directly reaches an external action. It may be logic-gated, where an IF or Switch node mediates the action. It may be human-gated, where an approval step is required before external side effects occur. It may also be low-autonomy, where the LLM only drafts, summarizes, or transforms text without controlling downstream action.

This motivates our RQ4. Existing agent benchmarks evaluate whether agents can act autonomously, but they do not usually measure how real workflow authors balance autonomy and human control in public automation templates \cite{liu2024agentbench,zhou2024webarena,xie2024osworld}. Our work fills this gap by classifying workflow designs according to how LLM outputs connect to control nodes, external action nodes, and human review steps. This is especially important because autonomy in workflow systems is not binary. For example, a workflow where an LLM drafts a Gmail response but stores it as a draft is different from a workflow where an LLM directly sends the email. The former preserves human control, while the latter delegates external action to the automation system.

\subsection{Summary and Positioning}

Prior work has made substantial progress on LLM reasoning, tool use, agent architectures, workflow generation, and benchmark-based evaluation \cite{yao2023react,schick2023toolformer,qin2024toolllm,wu2024autogen,fan2024workflowllm,liu2024agentbench,zhou2024webarena,jimenez2024swebench,xie2024osworld}. 
However, these studies primarily focus on model capabilities, agent frameworks, or researcher-defined benchmarks. They provide limited empirical evidence about how LLM agents are actually assembled in public low-code workflow ecosystems. Our work addresses this gap through a large-scale empirical study of n8n workflows. Specifically, we characterize what tasks LLMs are used for, how LLMs are structurally embedded with tools and services, what reliability mechanisms are encoded, and how workflows balance autonomous execution with human control.

\section{Conclusion}
Large Language Models are increasingly deployed as components of workflow-based automation systems, where their behaviour is shaped not only by the underlying model but also by workflow structure, external tools, control logic, reliability mechanisms, and human oversight. In this paper, we presented the first large-scale empirical study of LLM-based workflows in the N8n ecosystem. By analysing 6,003 publicly available workflows, we investigated the tasks performed by LLMs, the structural and tool-use patterns adopted by workflow authors, the reliability mechanisms encoded in workflows, and the extent to which workflows are designed as autonomous versus human-controlled systems. Our results show that LLMs are primarily used for text generation, information extraction, and agentic execution tasks. From an architectural perspective, LLMs are commonly embedded as intermediate processing components within larger automation pipelines rather than operating as standalone endpoints. We further observe growing adoption of agent-oriented compositions involving tools, memory, structured output parsers, and retrieval components. From a reliability perspective, many workflows rely on lightweight local error handling settings, while explicit end-to-end recovery mechanisms remain relatively uncommon. Finally, we find that most workflows favour automation over human oversight, with explicit human approval appearing in only a small fraction of workflows.

These findings provide an empirical view of how LLM agents are currently engineered in real-world automation environments. We believe that the dataset, methodology, and observations presented in this work can serve as a foundation for future research on workflow-aware agent evaluation, and  trustworthy LLM-based automation systems.

\newpage
\section*{Declarations}

\subsection*{Funding}
Zhou's work is partially supported by the National Natural Science Foundation of China (62572226).

\subsection*{Ethical approval}
Not applicable.

\subsection*{Informed consent}
Not applicable.

\subsection*{Author Contributions}
Yutian Tang contributed to conceptualization, methodology, data collection, data analysis, and manuscript writing. Yuming Zhou contributed to methodology, implementation, data analysis, and manuscript revision. Huaming Chen contributed to methodology, interpretation of results, and manuscript revision. All authors read and approved the final manuscript.

\subsection*{Data Availability Statement}
The dataset used in this study is derived from publicly available n8n workflow templates. To support reproducibility, we will release the data collection scripts, filtering scripts, analysis scripts, workflow identifiers, derived metadata, and aggregated result files in an archival repository upon acceptance. The released artifacts can be found: \\

\url{https://sites.google.com/view/n8n-empirical-study/home}

\subsection*{Conflict of Interest}
The authors declare that they have no conflict of interest.

\newpage

%
%



\bibliographystyle{spmpsci} 
\bibliography{reference.bib}

\end{document}